\pdfoutput=1

\documentclass[11pt]{article}

\usepackage{acl}

\usepackage{times}
\usepackage{latexsym}

\usepackage[T1]{fontenc}

\usepackage[utf8]{inputenc}
\usepackage{graphicx}
\usepackage{arydshln}
\usepackage{microtype}
\usepackage{inconsolata}

\usepackage{tabularx}
\usepackage{color,colortbl}
\usepackage[most]{tcolorbox}

\usepackage{xcolor}
\usepackage{adjustbox}
\definecolor{best}{RGB}{255, 159, 69}
\definecolor{gray}{RGB}{221, 221, 221}

\usepackage{booktabs}
\usepackage{multirow}
\usepackage{utfsym}
\usepackage{makecell}
\usepackage{soul}

\title{All Data on the Table: Novel Dataset and Benchmark for Cross-Modality Scientific Information Extraction}

\author{\bf Yuhan Li$^{\alpha}$\footnotemark[1] \footnotemark[2] \quad Jian Wu$^{\beta}$\footnotemark[1] \footnotemark[2] \quad Zhiwei Yu$^{\gamma}$ \quad Börje F. Karlsson$^{\delta}$\footnotemark[1] \\ 
\bf Wei Shen$^{\alpha}$ \quad Manabu Okumura$^{\beta}$ \quad Chin-Yew Lin$^{\gamma}$ \\
$^{\alpha}$Nankai University, $^{\beta}$Tokyo Institute of Technology, \\ $^{\gamma}$Microsoft Research Asia,  
$^{\delta}$Beijing Academy of Artificial Intelligence (BAAI)}

\begin{document}

\maketitle

\begin{abstract}
\renewcommand{\thefootnote}{\fnsymbol{footnote}}
\footnotetext[1]{This work was partially performed while the authors were interns or researchers at Microsoft Research Asia.}
\footnotetext[2]{Equal Contributions.}

Extracting key information from scientific papers has the potential to help researchers work more efficiently and accelerate the pace of scientific progress. Over the last few years, research on Scientific Information Extraction (SciIE) witnessed the release of 
several new systems and benchmarks. However, existing paper-focused datasets mostly focus only on specific parts of a manuscript (e.g., abstracts) and are single-modality (i.e., text- or table-only), due to complex processing and expensive annotations. Moreover, core information can be present in either text or tables or across both. To close this gap in data availability and enable cross-modality IE, while alleviating labeling costs, we propose a semi-supervised pipeline for annotating entities in text, as well as entities and relations in tables, in an iterative procedure. Based on this pipeline, we release \textsc{SciCM}, a high-quality scientific cross-modality IE benchmark, with a large-scale corpus and a semi-supervised annotation pipeline.
%
%
We further report the performance of state-of-the-art IE models on the proposed benchmark dataset, as a baseline\footnotemark[3]\footnotetext[3]{Our code and data will be available after acceptance.}. Lastly, we explore the potential capability of large language models such as ChatGPT for the current task. Our new dataset, results, and analysis validate the effectiveness and efficiency of our semi-supervised pipeline, and we discuss its remaining limitations.
\end{abstract}

\section{Introduction}
As scientific communities grow and evolve, there has been an explosion in the number of scientific papers being published in recent years\footnote{\url{https://arxiv.org/stats/monthly\_submissions}}, which makes it increasingly difficult for researchers to discover useful insights and new techniques in the vast amount of information. One approach to help researchers keep abreast with the latest scientific advances and quickly identify new challenges and opportunities is to automatically extract and organize crucial scientific information 
from collections of research publications \cite{viswanathan2021citationie}.
Scientific information extraction (SciIE) aims to extract such information from scientific literature corpora and has seen growing interest recently, with the rapid evolution of systems and benchmarks \cite{jain2020scirex,zhuang2022resel}. 

SciIE serves as an important pre-processing step for many downstream tasks, including scientific knowledge graph construction \cite{wang2021covid}, academic question answering \cite{dasigi2021dataset}, and method recommendation \cite{luan2018information}. Additionally, the development of scientific large language models (LLMs), such as Galactica \cite{taylor2022galactica} and Mozi \cite{science-llm}, allows the exploration of several practical science scenarios (e.g., suggest citations, ask scientific questions, and write scientific code). However, language models can hallucinate without verification. Moreover, language models are frequency-biased and often overconfident. SciIE along with appropriate QA or retrieval systems - e.g., TIARA \cite{shu-etal-2022-tiara} - can help alleviate such problems and facilitate model performance on downstream tasks, as similarly demonstrated in leveraging Wikidata to improve LLM factuality \cite{xu2023finetuned}.
\begin{table*}[]
\centering
\small
\begin{adjustbox}{width=2.05\columnwidth,center}
\begin{tabular}{llllc}
\toprule
\textbf{Benchmarks} & \textbf{IE Task} & \textbf{Modality \& Coverage} & \textbf{Domain} & \textbf{Size}  \\ \midrule
\textbf{\textsc{SemEval-2017 Task 10}} \cite{augenstein2017semeval} & NER, RE & Text (several paragraphs) & CS, MS, Phy & 500     \\
\textbf{\textsc{SemEval-2018 Task 7}} \cite{gabor2018semeval} & RE & Text (abstract) & CS (NLP) & 500    \\
\textbf{\textsc{SciERC}} \cite{luan2018multi} & NER, RE & Text (abstract) & CS & 500    \\
\textbf{\textsc{NLP-TDMS}} \cite{hou2019identification} & NER, ResE & Text (abstract), Tables & CS (NLP) & 332    \\
\textbf{\textsc{SciREX}} \cite{jain2020scirex} & NER, N-ary RE & Text (full) & CS (ML) & 438    \\
\textbf{\textsc{NLPContributions}} \cite{d2020nlpcontributions} & NER, RE & Text (several paragraphs) & CS (NLP) & 50    \\
\textbf{\textsc{TDMSci}} \cite{hou2021tdmsci} & NER & Text (several sentences) & CS (NLP) & 2,000    \\
\textbf{\textsc{ORKG-TDM}} \cite{kabongo2021automated} & NER, ResE & Text (several sections), Tables & CS & [5,361]    \\
\textbf{\textsc{TELIN}} \cite{yang2022telin} & NER, ResE & Tables & CS (ML) & 731    \\
\textbf{\textsc{GASP-NER}} \cite{otto2023gsap} & NER &  Text (full) & CS (ML) & 100    \\
\rowcolor{gray}\textbf{\textsc{SciCM}} (ours) & NER, ResE, RE & Text (full), Tables & CS, Stat, EESS, ... & 70 + [12,817] \\
\bottomrule
\end{tabular}
\end{adjustbox}
\vspace{-3mm}
\caption{An overview of existing scientific IE benchmarks. Domain acronyms: CS refers to Computer Science; MS refers to Material Science; Phy refers to Physics. Task acronyms: NER refers to Named Entity Recognition; RE refers to Relation Extraction; ResE refers to Result Extraction. ``[ ]'' indicates automated annotations.}
\label{tab:benchmarks}
\vspace{-4mm}
\end{table*}

Initial corpora and benchmarks extracted information from specific parts of a paper text, such as abstracts \cite{gabor2018semeval,luan2018multi} or selected paragraphs \cite{augenstein2017semeval,d2020nlpcontributions,hou2021tdmsci}. 
However, scientific entities are spread through the whole paper body; thus neglecting any text fragments or tables will likely result in missing key information. This is especially true in tables that condense complex information and data on experimental results.
\citet{jain2020scirex} first attempts to create a scientific IE benchmark at the document level and \citet{hou2019identification} first annotates entities in tables as well as image captions. 
Unfortunately, these benchmarks mostly focus on a single modality of paper content and coarser annotations due to the large gap between modalities, plus high processing and annotation costs. Text-only modality covers information presented in an unstructured and narrative way, whereas table-only modality for structured and concise information representation. Nonetheless, it is increasingly clear that information obtained from a single modality will inevitably miss critical information in a given paper, e.g., it is hard to extract experiment results and settings only from text or to extract models and metrics from tables. Combining modalities is thus critical and enables further scenarios like hybrid QA \cite{wu-etal-2023-tacr}. 
However, it is non-trivial to directly extract entities from tables due to the lack of context information and the label imbalance issue. Leveraging entities consistently appearing in both text and tables can intuitively facilitate this process.
To address the above-mentioned gaps in data availability, a new cross-modality and document-level benchmark dataset for SciIE is needed.

Annotating such a benchmark remains challenging because 1) it requires domain expertise and considerable annotation effort to comprehensively label a sizeable benchmark, 2) it requires annotators to understand the domain and the whole paper to maintain annotation consistency across different modalities, and 3) annotations need to be fine-grained enough (and consistent across documents) to unlock relevant semantics. To overcome these annotation challenges, we develop a semi-supervised pipeline for annotating entities in text and both entities and relations in tables of academic papers, which involves a two-stage iterative procedure. Specifically, a replaceable extractor is first trained on a small amount of high-quality manually annotated papers and then utilized to label a large number of papers automatically. Experts are introduced next to correct any false labels. This process can be repeated iteratively multiple times, with the extractor becoming more accurate as it can use the newly labeled data to improve its performance. During training, we also adopt label mapping in text extraction via leveraging existing benchmarks \cite{luan2018multi,jain2020scirex} to enrich our annotations. We also prove that informing the table extractor with text extraction results leads to more accurate and context-rich table annotations. 


Based on this pipeline, we release 1) \textsc{\textbf{SciCM}}, a high-quality expert annotated benchmark that supports multiple \underline{\textbf{Sci}}entific \underline{\textbf{C}}ross-\underline{\textbf{M}}odality IE tasks; 2) a large-scale corpus containing automatically annotated papers with different domains; 3) a visualization tool that enables researchers to get a global view of key information in scientific papers; and 4) the pipeline itself, which can be utilized as provided or further extended. We conduct experiments on \textsc{SciCM} utilizing SoTA IE models and popular LLMs such as ChatGPT to showcase its characteristics and as a baseline. To further investigate the efficacy of our pipeline, we assess both the quality and the speed of annotation on \textsc{SciCM}. Additionally, a thorough error analysis is conducted.

We summarize our key contributions as follows: 1) we are the first to explore cross-modality IE to build a bridge between text and tables in scientific articles; 2) we develop a semi-supervised pipeline for annotating scientific terms along with their relations without requesting lots of human supervision at document-level; 3) we release a high-quality benchmark that enables diverse 
SciIE tasks and a large-scale corpus to facilitate research over the scientific literature; 4) extensive experiments validate the efficiency, effectiveness, and adaptability of our semi-supervised pipeline.

\section{Related Work}

An overview of existing SciIE benchmarks is shown in Table \ref{tab:benchmarks}. The field of SciIE began with extracting information from only the text modality. \citet{augenstein2017semeval} proposed \textsc{SemEval-2017 Task 10} to support the task of identifying entities and relations in a corpus of $500$ paragraphs taken from open-access journals. \citet{gabor2018semeval} presented \textsc{SemEval-2018 Task 7} on relation extraction from $500$ abstracts of NLP papers. \citet{luan2018multi} released \textsc{SciERC} containing $500$ abstracts with more fine-grained types of entities (i.e., 6 types) and relations (i.e., 7 types). More recently, \citet{d2020nlpcontributions} and \citet{hou2021tdmsci} proposed benchmarks which are composed of several paragraphs from NLP papers, aiming to capture contributed scientific terms and extract <\texttt{Task}, \texttt{Dataset}, \texttt{Metric}> (TDM) triples, respectively. Since scientific terms can appear anywhere in the paper, \citet{jain2020scirex} proposed \textsc{SciREX} that tries to comprehensively annotate the full paper text. \citet{otto2023gsap} manually annotated \textsc{GASP-NER} for identifying named entities associated with the interplay between machine learning model entities and dataset entities. Inspired by them, we consider document-level extraction and define more complete entity types compared with prior works. We also leverage two high-quality benchmarks, i.e., \textsc{SciERC} and \textsc{SciREX}, to boost our text extractor for a subset of entities.

Extracting information from the table modality has attracted much research attention in recent years since they contain relevant structural knowledge that aids extraction. To the best of our knowledge, there are currently only three datasets that explore cross-modality extraction in the scientific domain: \textsc{NLP-TDMS} \cite{hou2019identification}, \textsc{ORKG-TDM} \cite{kabongo2021automated}, and \textsc{TELIN} \cite{yang2022telin}. These datasets associate score entities extracted from result tables with their corresponding TDM triples, but they only concentrate on simple result extraction and ignore many valuable types of entities that tables can provide such as \texttt{Model} and \texttt{Metric}. Additionally, there has been a lack of benchmarks supporting relation extraction in the table modality, which may lead to the absence of important relationships, as well as table-text relationships. According to these limitations, our goal is to develop a benchmark covering information extraction in both text and table modalities and across them, to close the gap in scientific data availability and facilitate more comprehensive and accurate information extraction.

\section{Preliminaries}


\subsection{Data Collection}


Extracting large amounts of tables, body text, and other metadata from scientific articles requires expert knowledge and comprehension of the article, which can be very time-consuming for expert annotation. To collect paper data,
we propose a method to automatically download and pre-process \LaTeX\ source files of scientific articles. Leveraging the structure of \LaTeX\ source files, we can easily locate each component of papers such as titles, sections, and tables. 


Specifically, we first search and crawl the arXiv \LaTeX\ sources from its official website\footnote{\url{https://arxiv.org/}} using its official library\footnote{\url{https://github.com/lukasschwab/arxiv.py}}. We design a simple parser to extract textual content (including all sections and sub-sections) and tables from a scientific paper. The titles of sections and sub-sections are retained. Each extracted table contains its caption and all table cells. To ensure that the extracted data is in a readable format for annotations, we utilize two public libraries, e.g., latexml\footnote{\url{https://math.nist.gov/~BMiller/LaTeXML/}} and dashtable\footnote{\url{https://dashtable.readthedocs.io/en/latest}} to ``clean" the extracted data. This process enables us to efficiently collect and pre-process large amounts of scientific papers for annotation and analysis.

\begin{figure*}[t]
  \centering
  \includegraphics[width=0.80\textwidth]{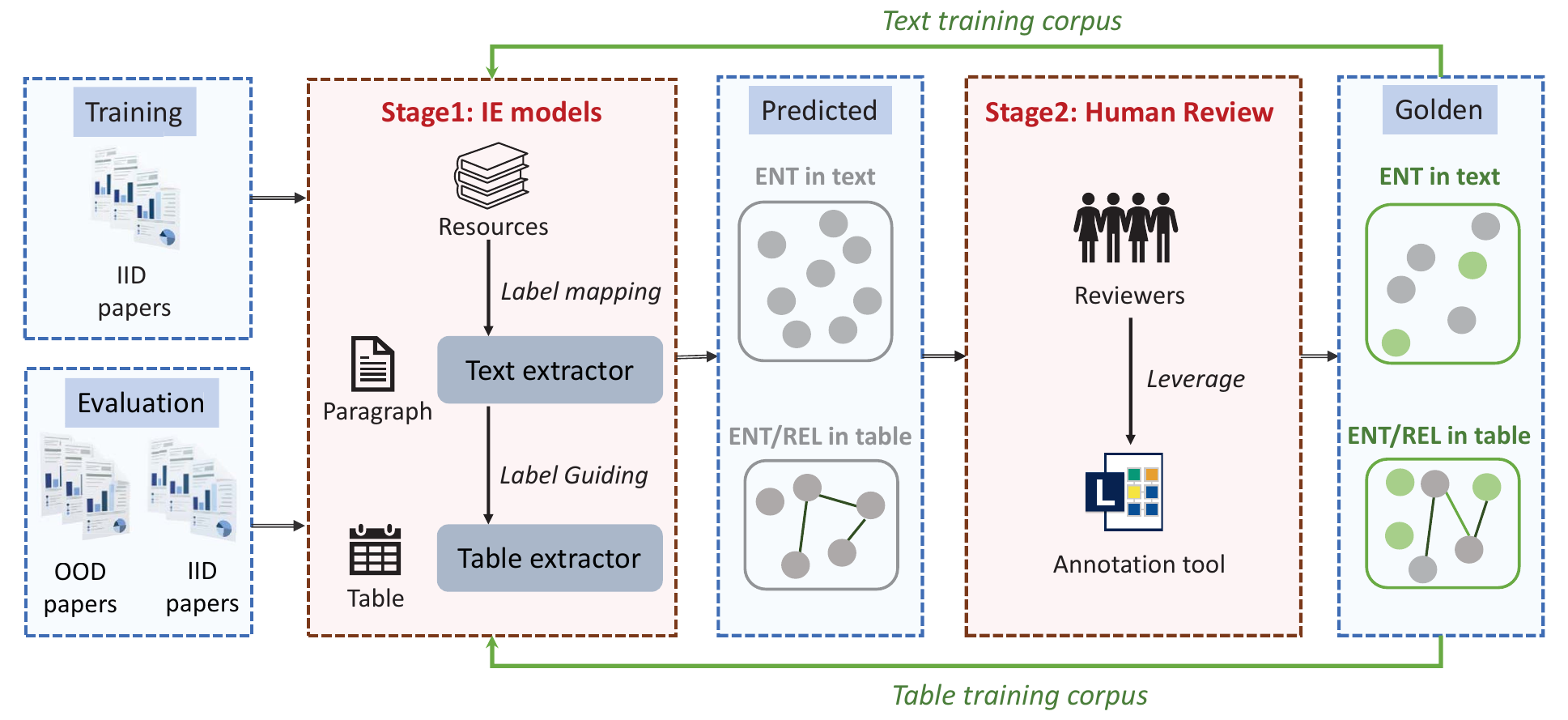}
  \caption{Overview of our semi-supervised pipeline. It consists of two stages: 1) training of text and table extractors through a limited number of manually annotated in-domain papers. Extractors are then employed to automatically annotate a larger set of both out-of-domain and in-domain papers; 2) expert reviewers rectify any false labels to obtain golden annotations, which are used to expand training and update extractors via iterative learning.}
  \label{fig:overview}
  \vspace{-5mm}
\end{figure*}

\subsection{Task Definition}

We aim to perform NER on a corpus of scientific papers across both text and table modalities. Unlike previous benchmarks \cite{augenstein2017semeval,luan2018multi}, we only perform RE on tables. 
Cross-sentence relations in text can be very challenging to be annotated \cite{jain2020scirex} and error prone. Formally, we denote each scientific paper as $D$, containing a sequence of paragraphs $P = \{p_1, p_2, ..., p_{|P|}\}$ and a sequence of tables $T = \{t_1, t_2, ..., t_{|T|}\}$. Each paragraph $p$ is composed of a sequence of sentences $\{s_1, s_2, ...s_n\}$ and each sentence is composed of a sequence of words $\{w_1, w_2, ..., w_i\}$. Each table $t$ is flattened via inserting separators and concatenated with its caption as a sequence of cells $\{c_{1,1}, c_{1,2}, .., c_{y,z}\}$, where $y$ and $z$ are the numbers of table rows and columns. Each cell is also composed of a sequence of words $\{w_1, w_2, ..., w_j\}$. Formally, the three IE tasks we support are defined as follows.

\paragraph{Text/Table NER} The goal of text/table NER is to examine all possible spans of words, denoted as $\{w_l, ..., w_r\}$ within a sentence/cell, where $l$ and $r$ represent the left and right indices of the span, and recognize if the span describes an entity and classifies it with its type if any.


\paragraph{Table RE} The goal of this task is to examine all unordered pairs of entities appearing within a given table, denoted as $(e_1, e_2)$, and determine the existence of any relationship between them.




\label{sec:semi-supervised-pip}



\section{Semi-supervised Annotation Pipeline}

\subsection{Overview}

The overview of our semi-supervised pipeline is depicted in Figure \ref{fig:overview}, which involves a two-stage iterative process. Specifically, two extractors are trained on a small amount of manually annotated or reviewed papers and leveraged to make automatic annotations on the unlabeled papers (Stage 1). Expert reviewers are introduced later to perform necessary corrections to generate high-quality annotations, which are utilized to update the extractors for iterative training (Stage 2). The pipeline can be repeated multiple times, with extractors becoming more accurate as they have the ability to use the newly labeled data to improve their extraction performance. 


Our proposed pipeline exhibits exceptional performance in the annotation of cross-modality entities and relations in the scientific literature. Specifically, it demonstrates the following advantages: i) \textbf{Efficiency}. The curation of datasets composed of long scientific documents with both paragraphs and tables can be a costly and labor-intensive task. However, our pipeline only requires a smaller number of labeled papers for training than fully-supervised learning, making it a cost-effective solution in comparison to existing benchmarks. ii) \textbf{Effectiveness}. We employ techniques such as label mapping and label guiding for text and tables respectively, to ensure high-quality annotations. iii) \textbf{Adaptability}. We conduct both in-domain (IID) and out-of-domain (OOD) evaluations to demonstrate that the pipeline can adapt to new domains with ease, which is particularly useful in the scientific domain where new research is continually being published. The performance of our pipeline will be further investigated in Section \ref{sec:experiments}.

\subsection{Text Information Extraction}

\paragraph{Entity types} We find that existing benchmarks \cite{luan2018multi,hou2019identification,jain2020scirex,hou2021tdmsci,kabongo2021automated} normally consider fine-grained knowledge with multiple entity types. However, all these datasets simply regard the ``models'' proposed by papers as ``methods'' and do not treat \texttt{Model} as a separate type of entity. It is not reasonable, as ``models'' and ``methods'' are distinct entities with different characteristics, with ``models'' being more representative of the paper and more likely to appear in experimental tables. To address this issue, we present more appropriate entity types, including \texttt{Task}, \texttt{Dataset}, \texttt{Metric}, \texttt{Model}, and \texttt{Method} (see Appendix \ref{subapp:entity}). 


\paragraph{Label mapping} Due to the mappable label sets between SciIE benchmarks, we are able to reuse some high-quality labels collected from existing benchmarks to boost the performance of extraction. Inspired by \cite{yang2017transfer,francis2019transfer}, we perform a label mapping step prior to training, using two fully-supervised annotated datasets  \cite{luan2018multi,jain2020scirex}. To be specific, we collect \texttt{Task}, \texttt{Metric}, and \texttt{Method} entities from \textsc{SciERC} \cite{luan2018multi} and \texttt{Task}, \texttt{Dataset}, \texttt{Metric}, and \texttt{Method} entities from \textsc{SciREX} \cite{jain2020scirex} to enrich our entity annotations. 

\paragraph{Extractor}
We apply the publicly available SciIE model PL-Marker \cite{ye2022packed} to perform text NER. PL-Marker inserts levitated markers in text and incorporate two packing strategies to achieve state-of-the-art F$_1$ scores and a notable level of efficiency on \textsc{SciERC} \cite{luan2018multi}. 

\subsection{Table Information Extraction}


\paragraph{Entity and relation types}

In contrast to text NER, in table NER we extract two additional types of entities:  \texttt{Score} and \texttt{Setting}. Scores rarely appear in the paper text but frequently appear in tables. We follow previous benchmarks \cite{hou2019identification,kabongo2021automated,yang2022telin} in extracting \texttt{Score} entities from tables to enable a more comprehensive understanding of the experimental results reported in scientific papers. \texttt{Setting} is another important entity type, as it refers to the context or environment in which the study was conducted. We use \texttt{Setting} entities to indicate different experimental settings, for example, BERT$_{large}$ and BERT$_{base}$, one-hop and multi-hop.


Similar to  \cite{hou2019identification,jain2020scirex,d2020nlpcontributions,hou2021tdmsci,kabongo2021automated}, we do not pre-define specific relation types and instead aim to uncover indirect relationships between pairs of entities. For more details please see Appendix \ref{subapp:relation}.

\paragraph{Label Guiding}

Inconsistency in NER predictions between tables and text can result in entities with the same name being assigned different entity types across different modalities. To address this issue, we utilize a label guiding rule that leverages the annotation results of the previous stage, i.e., text NER. Specifically, when an entity appears in both text and tables, we maintain consistency by assigning the same entity type in the table as the entity type identified in the text. This approach ensures that the entity types are consistent across different modalities.

\begin{figure}[ht]
    \centering
     \includegraphics[width=0.4\textwidth]{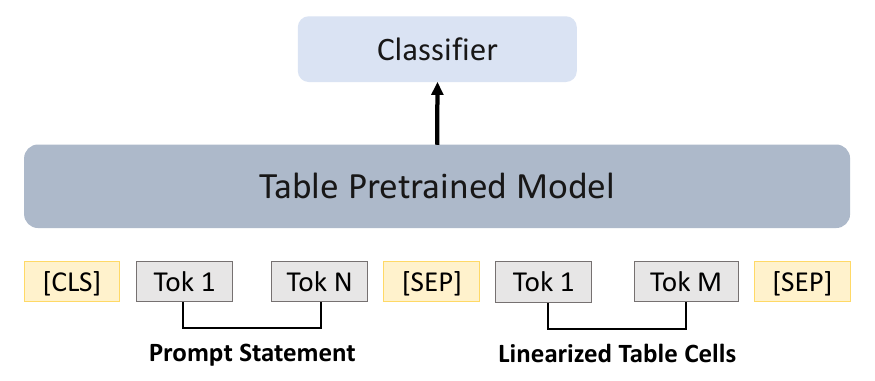}
    \caption{Table IE model architecture. A pre-trained model encodes the input table and its output is fed into a classifier to output the probabilities of entity types for each cell, or the relation probabilities of any two cells.} 
    \label{fig:table IE model}
    \vspace{-3mm}
\end{figure}

\paragraph{Table IE}
We treat table IE tasks, which include table NER and table RE, as classification tasks. The model structure of our table IE model is depicted in Figure \ref{fig:table IE model}, which comprises an encoding layer and a classification layer. 

Prompting the table NER model is straightforward. We prompt the model with a statement that specifies the entity type of a given cell in a table, i.e., ``The entity type of the cell $c_{i,j}$, in row $i$, column $j$ is $[E]$.'', where $[E] \in \Gamma$ and $\Gamma$ is the set of predefined entity types. For each table cell, we generate $m$ prompts and match these prompts with their entity types, where $m$ is the number of entity types. Let $E = \{ s_{i}, t_{i}, E^{+}_{i}, E^{-}_{i,1}, E^{-}_{i,m-1} \}^{m}_{i=1}$ be the training data that consists of $m$ instances. Each instance consists of a statement $s_{i}$ , a table $t_{i}$, a correct (positive) entity type $E^{+}_{i}$ , along with ($m$-1) wrong (negative) entity types $E^{-}_{i,j}$.

While TURL \cite{Deng2020TURLTU} can solve column relation extraction, it cannot extract relations in fine-grained granularity, such as at the cell level. To address this limitation, we design a solution for table RE, which is a binary classification task that determines whether two cells in a table have a relation or not. The template statement is ``Whether the cell $c_{i,j}$, located in row $i$, column $j$, has a relation with the cell $c_{p,q}$, located in row $p$, column $q$, or not?'', where the label is either $0$ or $1$. We follow the model architecture in Figure \ref{fig:table IE model}, where the input table is encoded by a pre-trained table model and a binary classifier is used to output the relation probabilities of any two cells in the same table.

\subsection{Visualization}

We utilize OneLabeler~\cite{Zhang2022OneLabelerAF}~\footnote{\url{https://microsoft.github.io/OneLabeler-doc}}, a tool to aid in the annotation process, enabling reviewers to label data objects of various entities and relations. To incorporate semi-supervised pipeline in the labeling tool, we start with the labeling workflow template and add a labeling module, which we configure to be implemented with the built-in tagger of OneLabeler. As a result, we are able to load the automatically extracted entities and relations into the tool, which allows annotators to just remove incorrect detections and create any missing false negative spans/relations, saving time and effort. More details about annotation rules and notes can be found in Appendix \ref{subapp:rulesandnotes}.


In Figure \ref{context annotation}, we show a portion of the text of the BERT \cite{Devlin2019BERTPO} paper, a milestone in the field of NLP with high citations. We annotate various types of entities, such as \texttt{Model}, \texttt{Task}, and \texttt{Method} in the abstract and other sections of the paper using different colors to distinguish between them, which allows annotators to efficiently and accurately label different kinds of entities. Due to space limitations, we provide two additional visualization examples in Appendix \ref{app:visualization}.

\begin{figure}[ht]
    \centering
    \includegraphics[width=0.5\textwidth]{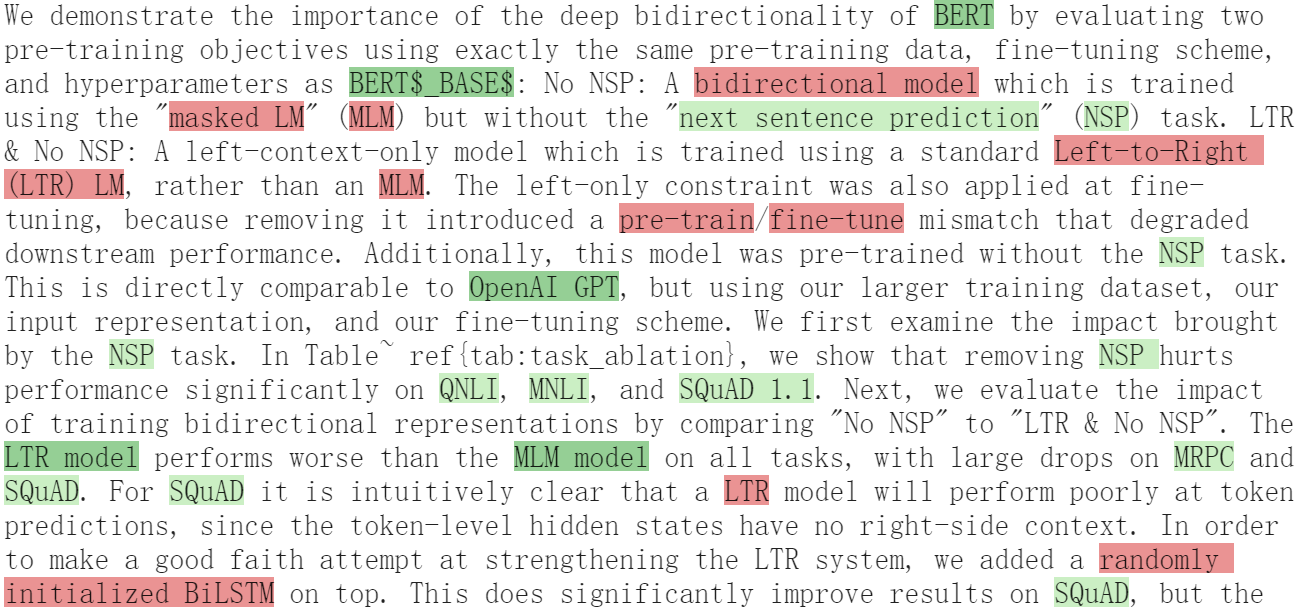}
    \caption{Visualization of annotated entities of BERT paper. Different types of entities are assigned distinct colors to facilitate their identification, with dark green representing the \texttt{Model}, light green representing the \texttt{Task}, and red representing the \texttt{Method}.}
    \centering
    \label{context annotation}
    \vspace{-6mm}
\end{figure}

\subsection{Released Benchmark}
This section describes the corpora constructed to train and evaluate our semi-supervised pipeline for cross-modality SciIE, as shown in Table \ref{tab:annotation} and the final dataset (SciCM).

Over 90\% of papers with arXiv IDs have source code available, allowing us to directly download and preprocess a large number of papers. For the first round of training, we use $10$ manually annotated papers as seeds in the computer science (CS) domain with human-annotated entities and relations. We then add $30$ new papers in the same domain and ask domain experts to review the model's predicted results on these papers. We use these human-reviewed $30$ papers as added corpus to further train a new version of extractors. We then automatically annotate other $30$ papers and review them for both in-domain (IID) and out-of-domain (OOD) evaluation, with $10$ papers each from CS, statistics (STAT), and electrical engineering and systems science (EESS), respectively. Finally, we utilize the extractors to automatically annotate $12817$ new papers in five different domains to facilitate research in scientific literature. This large-scale annotation of papers enables researchers to efficiently search for relevant information and extract insights from a large corpus of scientific literature.

\begin{table}[htbp]
  \centering
  \scalebox{0.75}{
    \begin{tabular}{lcccc}
    \toprule
          & \multicolumn{1}{c}{\textbf{Size}} &
          \multicolumn{1}{c}{\textbf{Domains}} &\multicolumn{1}{c}{\textbf{Auto}} & \multicolumn{1}{c}{\textbf{Reviewed}} \\
    \midrule
    \textbf{Seeds} &   10   &   CS  &  &  \usym{2714} \\
    \midrule
    \textbf{Added corpus} &   30    &   CS    &  \usym{2714} & \usym{2714} \\
    \midrule
    \textbf{Test set} &   30    &  CS/STAT/EESS     & \usym{2714}  & \usym{2714}\\
    \midrule
    \textbf{Corpus} &  12817   &  \makecell{CS/STAT/EESS \\ PHYSICS/MATH}   &  \usym{2714}  & \\
    \bottomrule
    \end{tabular}}%
    \caption{Domain distribution and labeling methods of different parts of \textsc{SciCM}.}
  \label{tab:annotation}%
\end{table}%

\vspace{-5mm}

\begin{table*}[h]
    \centering
    \small
    \begin{tabular}{lcccccccccc}
     \toprule
  \multirow{2}{*}{\textbf{Test domains}} & \multirow{2}{*}{\textbf{Settings}} & \multicolumn{3}{c}{\textbf{Text NER}}  & \multicolumn{3}{c}{\textbf{Table NER}} &  \multicolumn{3}{c}{\textbf{Table RE}}\\
  \cmidrule(lr){3-5}\cmidrule(lr){6-8} \cmidrule(lr){9-11} 
 &  & P & R & F$_1$ & P & R & F$_1$ & P & R & F$_1$\\\midrule
   \multicolumn{11}{c}{\textit{Round 1 Evaluation}}\\\midrule
\rowcolor{gray}\textbf{Overall} (IID \& OOD) & Fine-tune & 67.4  &  39.5 & 49.8  & 49.3  &  10.6 &  17.5 &  4.2 & \textbf{47.2} & 7.7 \\
\quad \scriptsize{* CS (IID)} & - & \scriptsize{67.4}  & \scriptsize{31.9}  & \scriptsize{43.3}  & \scriptsize{51.0}  &  \scriptsize{10.6} & \scriptsize{17.6}  &  \scriptsize{5.3} & \scriptsize{47.2} & \scriptsize{9.5}\\
\quad \scriptsize{* STAT (OOD)} & - & \scriptsize{71.6} & \scriptsize{47.7} & \scriptsize{57.2}  &  \scriptsize{55.1} & \scriptsize{11.0}  &  \scriptsize{18.4} &  \scriptsize{2.9} & \scriptsize{38.3} & \scriptsize{5.5}\\
\quad \scriptsize{* EESS (OOD)} & - &  \scriptsize{60.7} & \scriptsize{37.9}  & \scriptsize{46.7} &  \scriptsize{38.1} & \scriptsize{10.0}  & \scriptsize{15.8}  &  \scriptsize{5.0} & \scriptsize{66.7} & \scriptsize{9.3}\\
\midrule
  \multicolumn{11}{c}{\textit{Round 2 Evaluation}}\\\midrule
\rowcolor{gray}\textbf{Overall} (IID \& OOD) & Fine-tune & \textbf{79.0}  & \textbf{75.8} &  \textbf{77.4}  & 46.7 & \textbf{83.0} & \textbf{59.8} & \textbf{83.1} & 28.4 & 42.3 \\
\rowcolor{gray} & & \textit{\color{red} \scriptsize{+11.6\%}} & \textit{\color{red} \scriptsize{+36.3\%}} & \textit{\color{red} \scriptsize{+27.6\%}} & \textit{\color{blue} \scriptsize{-2.6\%}} & \textit{\color{red} \scriptsize{+72.4\%}} & \textit{\color{red} \scriptsize{+42.3\%}} & \textit{\color{red} \scriptsize{+78.9\%}} & \textit{\color{blue} \scriptsize{-18.8\%}} & \textit{\color{red} \scriptsize{+34.6\%}} \\
\quad \scriptsize{* CS (IID)} & - & \scriptsize{79.4} & \scriptsize{65.5} & \scriptsize{71.8}  & \scriptsize{50.3} & \scriptsize{83.6} & \scriptsize{62.8} & \scriptsize{83.8} & \scriptsize{21.8} & \scriptsize{34.6}\\
\quad \scriptsize{* STAT (OOD)} & - & \scriptsize{83.9} & \scriptsize{87.4} & \scriptsize{85.6}  & \scriptsize{41.8} & \scriptsize{76.2} & \scriptsize{53.9} & \scriptsize{81.5} &  \scriptsize{33.8} & \scriptsize{47.7}\\
\quad \scriptsize{* EESS (OOD)} & - & \scriptsize{71.0}  & \scriptsize{72.8} & \scriptsize{71.9}  & \scriptsize{39.7} & \scriptsize{80.8} & \scriptsize{53.1} & \scriptsize{85.6} & \scriptsize{35.9} & \scriptsize{50.6}\\

\midrule
  \multicolumn{11}{c}{\textit{ChatGPT Evaluation }}\\\midrule
  \rowcolor{gray}\textbf{Overall} (IID \& OOD)  & 1-shot ICL & 24.0 & 45.2  & 31.4  & 53.1 & 16.8 & 25.5 & 73.5 & 32.8 & 45.3\\
  \rowcolor{gray}\textbf{Overall} (IID \& OOD) & 2-shot ICL & 30.4 & 38.8 & 34.1 & \textbf{57.4} & 24.7 & 34.5 & 76.2 & 34.7 & \textbf{47.7}\\

     \bottomrule
    \end{tabular}
    \caption{Evaluating SoTA SciIE models on the test set with different domains. We evaluate in two rounds: \textit{Round 1} that is only trained on a small amount of paper seeds and \textit{Round 2} that is boosted leveraging added corpus. We also report the performance of ChatGPT on the few-shot ICL setting, providing a different number of demonstrations.}
    \label{tab:evaluation}
\vspace{-4mm}
\end{table*}

\vspace{-2mm}
\section{Experiments}
\label{sec:experiments}

\subsection{Dataset and Annotation Statistics}

\begin{table}[h]
\centering
\scalebox{0.80}{
\begin{tabular}{lrr}
\toprule
    \textbf{Statistics} (avg per paper)   & \textbf{Added corpus}  & \textbf{Test set} \\\midrule
    Sentences & 269.9 & 279.0 \\
    Words & 6787.7 & 7872.1 \\
    Tables & 5.7 & 3.9\\
    Cells & 390.3 &  190.3\\
    Entities in text & 133.2 & 152.9 \\
    Entities in tables & 151.6 & 94.4 \\
    Relations in tables & 81.3 & 59.5\\ \bottomrule
\end{tabular}}
\caption{Statistics of the added corpus and test set.}
\label{tab:dataset_stat}
\vspace{-6mm}
\end{table}

We evaluate our semi-supervised pipeline on the test set that includes both IID evaluation and OOD evaluation. Table \ref{tab:dataset_stat} presents the data statistics of the added corpus and test set, both of which are first automatically annotated and then manually reviewed. It can be seen from the table that scientific tables contain a considerable number of entities and relations, demonstrating the necessity of information extraction from tables. In addition, the long length (i.e., averaging $7000$+ words per paper) and cross-modality attributes of scientific papers make full annotation challenging, prompting us to focus on a more efficient annotation pipeline.

\subsection{Evaluation Metrics}

We follow the standard evaluation protocol \cite{zhong2021frustratingly} and use precision, recall, and F$_1$ as evaluation metrics. For text NER and table NER, both entity boundary and type are required to be correctly predicted. For table RE, the boundaries of the subject entity and the object entity should be correctly identified.



\subsection{Implementation Details}
    
Both our text and table IE models are implemented using Pytorch version $1$.$13$ and the Huggingface’s Transformers \cite{wolf-etal-2020-transformers} library, running on an A$100$ GPU. Specifically, we adopt in-domain \textit{scibert-scivocab-uncased} \cite{beltagy2019scibert} encoder with $110$M parameters for PL-Marker and \textit{tapas-base} encoder with $110$M parameters for our table IE models. During training, we adopt AdamW \cite{loshchilov2018decoupled} with a learning rate of $2$e-$5$ ($5$e-$5$ for tables) and a batch size of $16$ ($32$ for tables). Since scientific papers can be very long, we leverage cross-sentence information \cite{luan2019general,ye2022packed} to extend each sentence by its neighbor sentences and set the maximum length as $512$, which is the input length limit for many transformer-based models. 

\vspace{-2mm}
\subsection{Overall Performance}


We report the performance of our pipeline in Table \ref{tab:evaluation} with two rounds: \textit{Round 1} has access to only a small set of seeds for training, and \textit{Round 2} extends the training set by incorporating expert-reviewed papers. Our discussion focuses on the effectiveness and adaptability of our pipeline, taking into account the evaluation results. We also provide an analysis of ChatGPT's performance on the benchmark.

\vspace{-2mm}
\paragraph{Effectiveness study.}
The experimental results clearly show that the performance of our pipeline in \textit{Round 2} significantly outperforms that of \textit{Round 1} in all three IE tasks, with improvements of $27.6$, $42.3$, and $34.6$ F$_1$ points, respectively. It is attributed to the availability of more golden annotations for training in the added corpus. Specifically, text NER achieves better results in both rounds compared to the other tasks, as our label mapping step provides a useful and sufficient amount of NER labels, especially when training papers are few. In addition, we could observe that text IE yields better performance than table IE, possibly due to the fact that current table IE models can only rely on structural information and cannot utilize contextual semantics effectively. We discuss the label imbalance issue of table NER in Appendix \ref{app:variations table ner}.

\vspace{-1mm}
\paragraph{Adaptability study.}

In addition to performing IID evaluation, we also report OOD evaluation results based on papers from the STAT and EESS domains. The results indicate that our pipeline is capable of achieving comparable performance (i.e., $\pm$ $13.7$, $\pm$ $9.7$, $\pm$ $16.0$ F$_1$ points in three tasks) when tested on other domains, demonstrating its ability to adapt to new domains and generalize to unseen data. It is particularly valuable in the scientific domain, where new research is continually being published. Surprisingly, STAT yields the best performance even compared with the CS domain, possibly due to the sparsity of entities and relations in statistics papers, as well as the low-quality labeling by reviewers.

\vspace{-1mm}
\paragraph{ChatGPT Evaluation.}

LLMs pre-trained on massive corpora, such as ChatGPT, have demonstrated impressive few-shot learning ability on many NLP tasks. We investigate ChatGPT’s capabilities on our benchmark in terms of the few-shot In-context Learning (ICL) setting. To construct few-shot ICL prompts, we design the prompt template carefully and select demonstrations from the training set. For more details see Appendix \ref{app:prompt}. We use the official API\footnote{\url{https://platform.openai.com/docs}} to generate all outputs from ChatGPT. To prevent the influence of dialogue history, we generate the response separately for each testing sample. We compare ChatGPT with our pipeline for all sub-tasks in Table \ref{tab:evaluation}. Comparing 2-shot ICL with 1-shot ICL, it can be seen that providing more demonstrations generally leads to improvements (i.e., +$2.7$, +$9.0$, +$2.4$ F$_1$ points, respectively). We also observe that ChatGPT still struggles to achieve comparable performance with traditional fine-tuned IE models, indicating a need for further exploration in improving the performance of LLMs in cross-modality SciIE.



\vspace{-1mm}
\subsection{Annotation Speed}
Table \ref{tab:speed} presents the inference speed of our pipeline and ChatGPT. Our automatic labeling approach enables us to process $49.9$ sentences per second in text NER and $33.0$/$4.6$ tables per second in table IE. In text NER, our pipeline outperforms ChatGPT by achieving $77.4$ F$_1$ points and a faster processing speed of $49.9$ tables/s. Similarly, in table NER, our pipeline achieves a higher F$_1$ score ($59.8$) and a faster processing speed of $33.0$ tables/s, while ChatGPT achieves a lower F$_1$ score of $49.8$ and a slower processing speed of $0.2$ tables/s. However, in table RE, ChatGPT achieves a better F$_1$ score of $47.7$ compared to our table RE model, yet it still has a slower processing speed of $0.2$ tables/s.


\begin{table}[!t]
\begin{center}
\scalebox{0.73}{
\begin{tabular}{l cc cc cc}
    \toprule
    \multirow{3}{*}{{\textbf{Model}}} & \multicolumn{2}{c}{\textbf{Text NER}} & \multicolumn{2}{c}{\textbf{Table NER}} & \multicolumn{2}{c}{\textbf{Table RE}} \\
    \cmidrule(lr){2-3} \cmidrule(lr){4-5} \cmidrule(lr){6-7} & M & Speed & M & Speed & M & Speed\\
     & (F$_1$) & (sent/s) & (F$_1$) & (table/s) & (F$_1$) & (table/s)\\
    \midrule
    ChatGPT & 34.1 & 18.6 & 49.8 & 0.2 & \textbf{47.7} & 0.2 \\
    Our pipeline & \textbf{77.4} & \textbf{49.9} & \textbf{59.8} & \textbf{33.0} & 42.3 & \textbf{4.6}\\
    \bottomrule
\end{tabular}
}
\caption{We compare our pipeline and ChatGPT in both accuracy and annotation speed. The accuracy is measured as the F$_1$ on the test set. The speed is measured on a single A100 GPU with a batch size of $32$.}
\label{tab:speed}
\end{center}
\vspace{-6mm}
\end{table}

\vspace{-1mm}
\subsection{Error Analysis}

To further explore the limitations of our pipeline, we conduct an error analysis during the auto-annotation process and categorize major errors in both text and table modalities as follows. We also analyze ChatGPT's errors in Appendix \ref{app:errorofChatGPT}.

\vspace{-1mm}
\paragraph{Text IE Errors}
i) \textbf{Over Annotation.} It occurs when terms that are too general to offer useful scientific information are labeled as entities. For instance, ``neural network'' and ``natural language processing'' are examples of scientific terms that are mistakenly labeled as \texttt{Model} entities.
ii) \textbf{Missing Abbreviation.} Abbreviations appear frequently in scientific papers to represent entities with long names. Abbreviations are usually missed during auto-annotation. For example, ``coupled multi-layer attention'' can be recognized successfully, while its abbreviation ``CMLA'' is missed.
iii)  \textbf{Nested Entity Annotation.} Some entities that contain nested entities, such as ``Bi-LSTM-CRF + CNN-char'', should be recognized as a complete entity. However, the extractor tends to extract ``Bi-LSTM-CRF'' and ``CNN-char'' separately, leading to incomplete annotation.
iv) \textbf{Inconsistent Annotation.} Same entity is specified with different entity types even if it appears in the same paragraph.
v) \textbf{Insensitive to Context.} Sometimes an entity will be recognized as different types due to the different contextual environments. 


\vspace{-1mm}
\paragraph{Table IE Errors}
    i) \textbf{Entity Type Error.}
    It occurs when an entity in the table is labeled with the wrong entity type.
    ii) \textbf{Inconsistency in Table Structure.} 
    The structure of a table always provides important context and cues for extraction. Sometimes, there may be obvious inconsistencies in the entity types within the same column or row. For instance, recognizing \texttt{Dataset}, \texttt{Model}, and \texttt{Score} entities in the same column or row.
    iii) \textbf{NER Misleads RE.}
    It occurs when the NER model incorrectly identifies an entity, leading to incorrect relationship extraction of the current entity.
    iv) \textbf{Missing Relations:}
    RE model fails to identify a relationship between cells. 

\vspace{-1mm}
\section{Conclusion and Future Work}
In this paper, we present \textsc{SciCM}, a novel dataset for training and evaluating cross-modality SciIE. Along with it, we propose a semi-supervised pipeline that automatically annotates entities and relations with high effectiveness and efficiency. Our pipeline performs well across SciIE tasks, with good inference speed for iterative runs. Moreover, we release a visualization tool that helps users to annotate scientific items with a global view. In future work, we plan to extend our corpus by incorporating images from scientific papers. We hope our release can facilitate downstream tasks in the scientific domain.



\section*{Limitations}

In this paper, our focus was on proposing a benchmark that includes paper text and tables and a semi-supervised pipeline for automatic annotation. However, we did not consider images in the papers, which also contain a wealth of information. Beyond tables, images often illustrate the work process, pipeline, or framework presented in the scientific paper. In our future work, we plan to introduce and process images from scientific papers into our benchmark to further assist researchers in comprehending papers. This will enable us to provide a more holistic approach to understanding scientific papers and ensure that the benchmark covers all important modalities of information in scientific papers.

\section*{Ethics Statement}
We collect paper from the free distribution service arXiv\footnote{\url{https://arxiv.org/}}. The random crawled papers collected may not have been peer-reviewed. Currently, extraction is based on the assumption of the correctness of the public papers.

\bibliography{anthology}
\bibliographystyle{acl_natbib}
\clearpage 

\appendix

\section{Annotation Guideline}
\label{app:annotation guideline}

\subsection{Entity Category}
\label{subapp:entity}

\begin{itemize}
\item \textbf{Task:} The specific task or problem that the paper aims to address. E.g., information extraction, machine reading comprehension, image segmentation, etc.

\item \textbf{Model:} A formal representation or abstraction of the proposed system, which can be applied to solve the specific \textbf{Task}. E.g., BERT, ResNet, etc.

\item \textbf{Method:} The approach, technique, and tool that is used to construct the \textbf{Model} to solve the \textbf{Task}. E.g., self-attention, data augmentation, Adam, batch normalization, etc. 

\item \textbf{Dataset:} A collection of data that is used for training, validating, and testing the proposed \textbf{Model}. E.g., GLUE, COCO, CoNLL-2003, etc.

\item \textbf{Metric:} A quantitative measure or evaluation criterion that is used to assess the performance or quality of a \textbf{Model}. E.g., accuracy, F$_1$ score, etc.

\item \textbf{Setting:} It often appears in tables and refers to the context or environment in which the study was conducted. \textbf{Setting} is an important aspect of a research paper, as it provides the necessary information to reproduce or replicate the experimental results. E.g., one-hop, multi-hop, dev set, test set, etc. 

\item \textbf{Score:} It refers to a numerical value that is used to evaluate the performance of a \textbf{Model} on a specific \textbf{Task}, using a particular \textbf{Dataset} and \textbf{Metric}, under a specific \textbf{Setting}.

\end{itemize}

\subsection{Relation Category}
\label{subapp:relation}

Due to the relative scarcity of explicit relationships in body text, we only focus on annotating relations presented in tables. Similar to previous SciIE benchmarks \cite{hou2019identification,jain2020scirex,d2020nlpcontributions,hou2021tdmsci,kabongo2021automated}, we link different types of entities that are related, without requiring a specific type. This methodology allows for a more flexible representation of the relationships between entities, enabling an easier and deeper understanding of the underlying patterns and connections within the table content. 

\subsection{Rules and Notes}
\label{subapp:rulesandnotes}

Considering that annotators may have varying understandings of the annotation details, we have defined a set of rules and notes to standardize the annotation process. The rules defined are the following:

\begin{enumerate}
    \item Differences between \textbf{Model} and \textbf{Method}: A \textbf{Model} entity refers to the name of a model that can be applied to a specific task independently, while a \textbf{Method} entity cannot be used directly to solve a problem or task but can help models improve their performance.

    \item The proposed framework, which stacks several models should be classified into \textbf{Model}. For example, ``YOLOv5'' should be annotated as \textbf{Model}.

    \item We should annotate a combination of a \textbf{Model} and a \textbf{Method} as a \textbf{Method} rather than as a \textbf{Model}. For example, the ``RNN-based encoder'' belongs to a \textbf{Method}.

    \item Terms such as ``networks", ``neural", and ``model", which do not convey specific meaning and often appear at the beginning or end of entity names, should be excluded. For example, ``FasterRCNN network'' $\rightarrow$ ``FasterRCNN'',  ``neural machine translation'' $\rightarrow$ ``machine translation'', and ``question answering model'' $\rightarrow$ ``question answering''

    \item Avoid annotating broad and unclear noun phrases as entities. Scientific terms should be specific and well-defined concepts to ensure clarity and precision. For example, phrases such as ``neural network" and ``encoder-decoder architecture" should not be considered as \textbf{Model} entities.

    \item Adjectives that are not directly related to the domain of the research, such as ``state-of-the-art", should be avoided. Instead, more specific adjectives that accurately describe the \textbf{Model} or \textbf{Method} should be kept. For instance, ``Bidirectional LSTM''.

    \item Do not include any determinators (e.g., ``the'', ``a''), or adjective pronouns (e.g., ``this'', ``its'', ``these'', ``such'') to the entity span.

    \item Two entities with the same entity type should not have a relationship.

\end{enumerate}

\section{Visualization Examples} 
\label{app:visualization}

\begin{figure}[h]
    \centering
    \includegraphics[width=0.5 \textwidth]{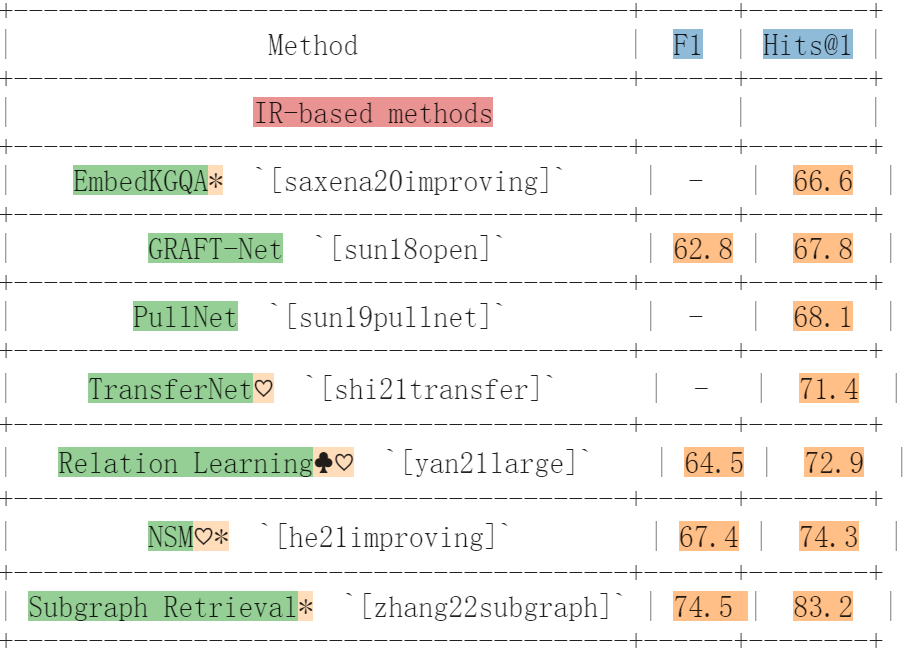}
    \caption{Visualization of annotated entities in tables.}
    \centering
    \label{tab:tablevisual}
\end{figure}

\noindent Figure \ref{tab:tablevisual} displays a visualization of Table 2 from the TIARA paper, which can be found on arXiv with id ``2210.12925''. We annotate various types of entities in tables using different colors to make them easily identifiable, which allows reviewers to efficiently and accurately label different kinds of entities and discover relations. Specifically, green represents the \texttt{Model}, dark orange represents the \texttt{Score}, light orange represents the \texttt{Setting}, blue represents \texttt{Metric}, and red represents the \texttt{Method}.

\begin{figure}[h]
    \centering
    \includegraphics[width=0.5\textwidth]{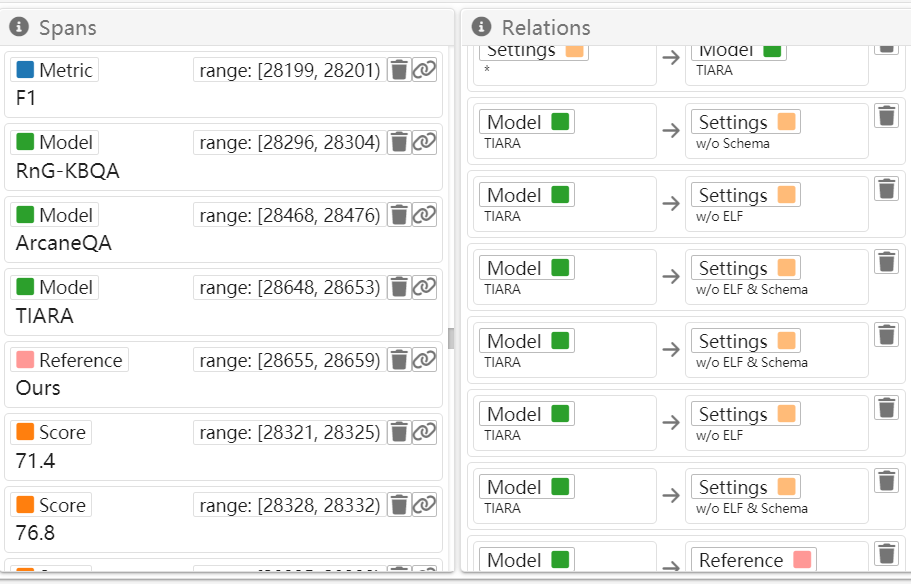}
    \caption{The interface for visualizing NER and RE annotations consists of two rectangles. The left rectangle, labeled ``Spans'', shows the entity name, entity type, and entity span. The right rectangle, labeled ``Relations'', shows the related entities within their types.}
    \centering
    \label{interface}
\end{figure}

The interface for annotating entities and relations is depicted in Figure \ref{interface}. Each entity is labeled with an associated entity type and its corresponding start and end positions within the document. Each relation is labeled with two entities. Entity annotation is done by directly selecting the boundary of the entity, while relation annotation is performed by clicking on the second button in the upper-right corner of the entity box. All entities and relations can be deleted by clicking the delete button.

 \begin{table}[t!]
    \small
    \centering
    \scalebox{0.9}{
    \begin{tabular}{p{1.8cm}p{6.2cm}}
    \toprule
    \multicolumn{2}{c}{\textit{\textcolor{blue}{Text NER}}}\\ \midrule
    \textbf{Error types} & \textbf{Cases}\\ \midrule
    Missing entities & ChatGPT struggles with extracting entities with long names. For example, ``Question-Answer driven Semantic Role Labeling'' fails to be extracted as a \textbf{Task} entity, and ``dual discourse-level target-guided strategy'' fails to be accurately identified as a \textbf{Method} entity. \\ \midrule 
    Undefined types & In text NER, even with a predefined set of entity types, it is common for some miscellaneous types to be introduced, such as [``Gábor et al.'', \textbf{Author}], [``Danqi Chen'', \textbf{Person}], etc. \\ \midrule \midrule
    \multicolumn{2}{c}{\textit{\textcolor{blue}{Table NER}}}\\ \midrule
    \textbf{Error types} & \textbf{Cases}\\ \midrule
    Unannotated entities & ``Our model'' in the table is not an entity but has been erroneously predicted as a \textbf{Model} entity. Similarly, ``Train data'' in the table is not an entity but has been incorrectly predicted as a \textbf{Dataset} entity.\\ 
    \midrule
    Incorrect types & ``K-means'' should be labeled as a \textbf{Method} entity, but it has been incorrectly predicted as a \textbf{Model} entity. Similarly, both ``dev set'' and ``test set'' should be labeled as \textbf{Setting} entities, but they have been incorrectly predicted as \textbf{Dataset} entities.\\
    \midrule \midrule

    \multicolumn{2}{c}{\textit{\textcolor{blue}{Table RE}}}\\ \midrule
    \textbf{Error types} & \textbf{Cases}\\ \midrule
    Missing relations & If a cell contains a \textbf{Model} entity, it should be annotated in relation to other cells in the same row that contain \textbf{Score} entities. However, this relationship is often missing in ChatGPT.\\ 

    \bottomrule
    \end{tabular}}
    \caption{Cases of different types of extraction errors that can be output by ChatGPT.}
    \label{tab:casechatgpt}
\end{table}

\section{Error Analysis of ChatGPT}
\label{app:errorofChatGPT}

In this section, we follow \cite{han2023information} to categorize and analyze ChatGPT’s errors on our three sub-tasks. Through manual checking, we find that the errors mainly include:

\begin{itemize}
    \item \textbf{Missing entities/relations: }  Missing one or more annotated target entities or relations.
    \item \textbf{Unannotated entities/relations: } Output the entities or relations that are not annotated in the test set.
    \item \textbf{Incorrect types: } The entity boundaries are correct, while the corresponding type comes from the set of pre-defined types, but does not match the annotated type.
    \item \textbf{Undefined types: } Output the types beyond the pre-defined types when the corresponding entity boundaries are correct.
    \item \textbf{Others: } Other errors apart from the above errors, such as incorrect output format, output unexpected information, etc.
\end{itemize}

Table \ref{tab:casechatgpt} presents several cases of extraction errors made by ChatGPT.
    
\section{Discussion of Table NER}
\label{app:variations table ner}

\begin{figure}[t]
    \centering
    \includegraphics[width=0.4\textwidth]{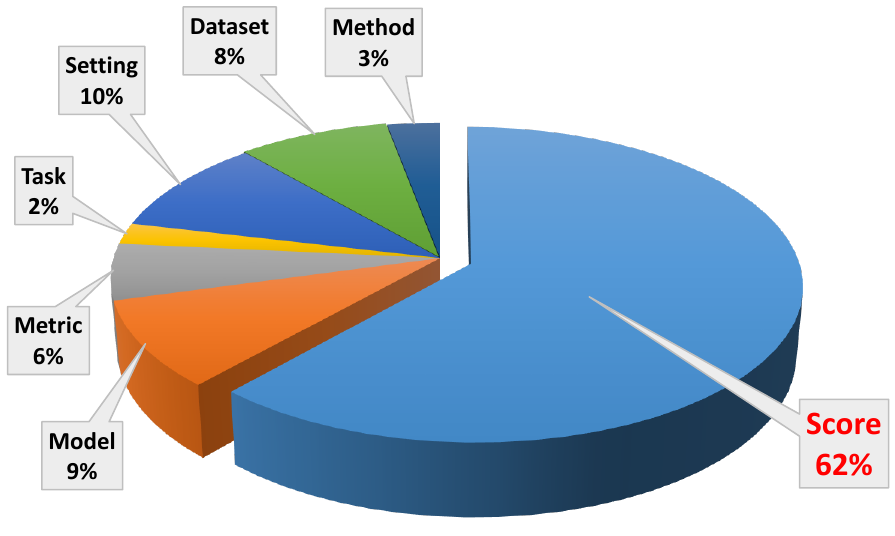}
    \caption{The entity types distribution of tables in the added corpus. }
    \centering
    \label{fig:entity_distribution}
    \end{figure}

As shown in Figure \ref{fig:entity_distribution}, we observe a long-tail problem of entity type distribution of tables, where \textbf{Score} entity occupies over 60\% of the total entity number. In this case, table NER models are more likely to predict the frequent \textbf{Score} entity, which can influence the model's generalization ability. Additionally, some numerical values in tables are simply parameters, data sizes, and other experimental information, but are prone to be predicted as \textbf{Score} entities. 

In text NER and table RE tasks, label imbalance is unlikely to occur since \textbf{Score} entities are relatively rare in text NER, and there is no specific type of relation in table RE. To address this label imbalance issue, we try to re-train a table NER model and test ChatGPT without the \textbf{Score} entities.

\subsection{Model Re-train}
We remove the \textbf{Score} entities, and the experiment results shown in Table \ref{tab:our model without score} demonstrate that our table NER without \textbf{Score} entities outperforms the previous table NER model, with improvements of $4.7$, $4.9$, and $5.1$ points on precision, recall, and F$_1$, respectively.

\begin{table}[t]
    \centering
    \small
    \begin{tabular}{lccc}
     \toprule
  \multirow{2}{*}{\textbf{Settings}} & \multicolumn{3}{c}{\textbf{Table NER}} \\\cmidrule(lr){2-4}
\cmidrule(lr){2-4} & P & R & F$_1$ \\\midrule
  Table NER w Scores & 46.7 & 83.0 & 59.8 \\
  Table NER w/o Scores & \textbf{51.4} & \textbf{87.9} & \textbf{64.9} \\
     \bottomrule
    \end{tabular}
    \caption{The performance of our table NER model with and without the \textbf{Score} entities.}
    \label{tab:our model without score}
\end{table}

\begin{table}[t]
    \centering
    \small
    \begin{tabular}{lccc}
     \toprule
  \multirow{2}{*}{\textbf{Settings}} & \multicolumn{3}{c}{\textbf{Table NER}} \\\cmidrule(lr){2-4}
\cmidrule(lr){2-4} & P & R & F$_1$ \\\midrule
  1-shot ICL w Scores & 53.1 & 16.8 & 25.5 \\
  1-shot ICL w/o Scores & 52.3 & 41.5  & 46.3 \\ \midrule
  2-shot ICL w Scores& \textbf{57.4} & 24.7 & 34.5 \\
  2-shot ICL w/o Scores & 53.2 & \textbf{49.2} & \textbf{51.1} \\
     \bottomrule
    \end{tabular}
    \caption{The performance of ChatGPT in the few-shot ICL setting with and without the \textbf{Score} entities.}
    \label{tab:chatgpt without score}
\end{table}

\subsection{ChatGPT without Scores}

We conduct experiments on ChatGPT without the \textbf{Score} entities, and as shown in Table \ref{tab:chatgpt without score}, the experimental results indicate that removing the \textbf{Score} entities leads to a significant improvement in recall for ChatGPT on both 1-shot and 2-shot ICL settings. However, it is important to note that removing the \textbf{Score} entities also results in a reduction in precision. Overall, our findings suggest that removing the \textbf{Score} entities could be a promising approach to improving the performance of table NER.

\section{Exemplar of the Prompt}
\label{app:prompt}

When evaluating large language models, prompting is a brittle process wherein small modifications to the prompt can cause large variations in the model predictions, and therefore significant effort should be dedicated to designing a painstakingly crafted perfect prompt for the given task \cite{arora2022ask,diao2023active}. In this study, we follow \cite{li2023evaluating} and \cite{han2023information}, who have conducted extensive evaluations of the power of ChatGPT in multiple information extraction tasks, to design prompt templates that are well-suited to our tasks. We investigate the performance of few-shot In-context Learning (ICL) on our benchmark. To eliminate the randomness, we manually select two demonstrations for each task, ensuring that all entity types are covered.

We give our designed input examples for three different kinds of scientific information extraction tasks to help readers understand our implementation, as shown in Table \ref{tab:textner}, \ref{tab:tablener}, and \ref{tab:tablere}, respectively. 

\begin{table*}[!t]\footnotesize
\centering
\small
\label{table:baseprompt}
\begin{tabular}{p{0.95\linewidth}}
\toprule
\textbf{\textit{Prompts of text NER (2-shot In-context Learning)}} \\ \midrule
\textbf{Scientific named entity extraction is a task in natural language processing that aims to identify specific entities with semantic meaning from paper text and classify them into predefined types.} \\
 \\ \\

\textbf{Paper text is typically segmented into sequences of tokens and each entity is labeled with a tag indicating its type. Entity types include:} \\ \\\\

\textbf{Task:} The specific task or problem that the paper aims to address. E.g., information extraction, machine reading comprehension, image segmentation, etc.

\textbf{Model:} A formal representation or abstraction of the proposed system, which can be applied to solve the specific \textbf{Task}. E.g., BERT, ResNet, etc.

\textbf{Method:} The approach, technique, and tool that is used to construct the \textbf{Model} to solve the \textbf{Task}. E.g., self-attention, data augmentation, Adam, batch normalization, etc. 

\textbf{Dataset:} A collection of data that is used for training, validating, and testing the proposed \textbf{Model}. E.g., GLUE, COCO, CoNLL-2003, etc.

\textbf{Metric:} A quantitative measure or evaluation criterion that is used to assess the performance or quality of a \textbf{Model}. E.g., accuracy, F$_1$ score, etc. \\\\\\

\textbf{Here are two demonstrations:} \\\\\\

\textbf{\textit{\textcolor{purple}{Given type set:}}} [Task, Model, Method, Dataset, Metric]. 

\textbf{\textit{\textcolor{red}{Sentences:}}} Our task is to classify images in the CIFAR-10 dataset into their respective classes, which include animals, vehicles, and household items. This task has practical applications in areas such as autonomous driving, object recognition, and image search. In this paper, we propose a novel approach for image classification using a deep learning model based on the EfficientNet architecture and transfer learning techniques. Our proposed model, named EfficientNet-Transfer, is a modified version of the EfficientNet-B0 architecture that has been pre-trained on the ImageNet dataset and fine-tuned on our target dataset. We use the CIFAR-10 dataset, which contains 60,000 32x32 pixel color images in 10 classes, as our target dataset. We evaluate our model using classification accuracy, precision, recall, and F1-score. 

\textbf{\textit{\textcolor{brown}{Question:}}} Please extract the named entity from the given sentences. Based on the given label set, provide the extraction results in the format: [[Entity Name, Entity Type]] without any additional things including your explanations or notes. If there is no entity in the given sentence, please return a null list like []. 

\textbf{\textit{\textcolor{cyan}{Entities:}}} [[`CIFAR-10', `Dataset'], [`autonomous driving', `Task'], [`object recognition', `Task'], [`image search', `Task'], [`image classification', `Task'], [`EfficientNet', `Method'], [`transfer learning', `Method'], [`EfficientNet-Transfer', `Model'], [`EfficientNet-B0', `Model'], [`ImageNet', `Dataset'], [`accuracy', `Metric'], [`precision', `Metric'], [`recall', `Metric'], [`F1-score', `Metric']] \\\\\\

\textbf{\textit{\textcolor{purple}{Given type set:}}} [Task, Model, Method, Dataset, Metric].

\textbf{\textit{\textcolor{red}{Sentences:}}} We perform sentiment analysis on movie reviews using deep learning techniques. We propose AttRNN, a novel model architecture based on a recurrent neural network (RNN) with attention mechanisms. Our model integrates both word-level and sentence-level attention mechanisms to improve its discriminative power and capture more relevant features from the input text. We evaluate our proposed model on the Movie Review Sentiment Analysis dataset, which consists of 50,000 movie reviews labeled as positive or negative. We use the accuracy metric to measure the performance of our model, which is defined as the percentage of correctly classified movie reviews in the test set. We also report the F1-score, which takes into account both precision and recall of the positive and negative classes.

\textbf{\textit{\textcolor{brown}{Question:}}} Please extract the named entity from the given sentences. Based on the given label set, provide the extraction results in the format: [[Entity Name, Entity Type]] without any additional things including your explanations or notes. If there is no entity in the given sentence, please return a null list like [].

\textbf{\textit{\textcolor{cyan}{Entities:}}} [[`sentiment analysis', `Task'], [`AttRNN', `Model'], [`recurrent neural network', `Method'], [`RNN', `Method'], [`word-level and sentence-level attention mechanisms', `Method'], [`attention mechanisms', `Method'], [`Movie Review', `Dataset'], [`accuracy', `Metric'], [`F1-score', `Metric']]  \\\\\\

\textbf{\textit{\textcolor{purple}{Given type set:}}} [Task, Model, Method, Dataset, Metric]. 

\textbf{\textit{\textcolor{red}{Sentences:}}} \textcolor{red}{[S]}

\textbf{\textit{\textcolor{brown}{Question:}}} Please extract the named entity from the given sentences. Based on the given label set, provide the extraction results in the format: [[Entity Name, Entity Type]] without any additional things including your explanations or notes. If there is no entity in the given sentence, please return a null list like [].

\textbf{\textit{\textcolor{cyan}{Entities:}}}  \\

\bottomrule
\caption{The prompt template of text NER leveraging 2-shot In-context Learning. \textcolor{red}{[S]} denotes the sentences we want to extract.} 

\label{tab:textner}
\end{tabular}
\end{table*}

\begin{table*}[!t]\footnotesize
\centering
\small
\label{table:table_NER_prompt}
\begin{tabular}{p{0.95\linewidth}}
\toprule
\textbf{\textit{Prompts of table NER (2-shot In-context Learning) without Score entities}} \\ \midrule
\textbf{Considering 6 entity types including `Task', `Model', 'Method', `Dataset', `Metric', `Setting'. 
} \\
 \\ \\

\textbf{Here are two demonstrations:} \\\\\\

\textbf{\textit{\textcolor{purple}{Given type set:}}} [`Task', `Model', `Method', `Dataset', `Metric', `Setting']. 

\textbf{\textit{\textcolor{red}{Table:}}} 
[[`System', `MNLI-(m/mm)', `QQP', `QNLI`, `SST-2', `CoLA', `STS-B', `MRPC', `RTE', `Average'], 
[`', `392k', `363k', `108k', `67k', `8.5k', `5.7k', `3.5k', `2.5k', `-'], 
[`Pre-OpenAI SOTA', `80.6/80.1', `66.1', `82.3', `93.2', `35.0', `81.0', `86.0', `61.7', `74.0'], [`BiLSTM+ELMo+Attn', `76.4/76.1', `64.8', `79.8', `90.4', `36.0', `73.3', `84.9', `56.8', `71.0'], [`OpenAI GPT', `82.1/81.4', `70.3', `87.4', `91.3', `45.4', `80.0', `82.3', `56.0', `75.1'],
[`bertbase', `84.6/83.4', `71.2', `90.5', `93.5', `52.1', `85.8', `88.9', `66.4', `79.6'], 
[`bertlarge', `86.7/85.9', `72.1', `92.7', `94.9', `60.5', `86.5', `89.3', `70.1', `82.1']]

\textbf{\textit{\textcolor{brown}{Question:}}} Please extract the named entity from the given table and output a JSON object that contains the following: \{`Task': [list of entities], `Dataset': [list of entities], `Model': [list of entities], `Method': [list of entities], `Metric': [list of entities], `Setting': [list of entities]\}. If no entities are presented in any categories keep it None.

\textbf{\textit{\textcolor{cyan}{Entities:}}} 
\{`Task': [`QQP', `MRPC'], `Dataset': [`MNLI-(m/mm)', `QQP', `QNLI', `SST-2', `CoLA', `STS-B', `MRPC ', `RTE', `Average', `GLUE Test', `WNLI set', `STS-B'], `Model': [`Pre-OpenAI', `BiLSTM+ELMo+Attn', `OpenAI GPT', `bertbase', `bertlarge', `BERT', `OpenAI GPT', `BERT'], `Method': [], `Metric': [`F1 scores', `Spearman correlations', `accuracy scores'], `Setting': []\}  \\\\\\

\textbf{\textit{\textcolor{purple}{Given type set:}}} [`Task', `Model', `Method', `Dataset', `Metric', `Setting']. 

\textbf{\textit{\textcolor{red}{Table:}}} 
[[`System', `Dev', `Dev', `Test', "Test'], [`', `EM', `F1', `EM', `F1'], [`Top Leaderboard Systems (Dec 10th, 2018)', `Top Leaderboard Systems (Dec 10th, 2018)', `Top Leaderboard Systems (Dec 10th, 2018)', `Top Leaderboard Systems (Dec 10th, 2018)', `Top Leaderboard Systems (Dec 10th, 2018)'], [`Human', `-', `-', `82.3', `91.2'], [`\#1 Ensemble - nlnet', `-', `-', `86.0', `91.7'], [`\#2 Ensemble - QANet', `-', `-', `84.5', "90.5'], [`Published', `Published', `Published', `Published', `Published'], [`BiDAF+ELMo (Single)', `-', "85.6', `-', `85.8'], [`R.M. Reader (Ensemble)', `81.2', `87.9', `82.3', `88.5'], [`Ours', `Ours', `Ours', `Ours', `Ours'], [`bertbase(Single)', `80.8', `88.5', `-', `-'], [`bertlarge(Single)', `84.1', `90.9', `-', `-'], [`bertlarge(Ensemble)', `85.8', `91.8', `-', `-'], [`bertlarge(Sgl.+TriviaQA)', `84.2', `91.1', `85.1', `91.8'], [`bertlarge(Ens.+TriviaQA)', `86.2', `92.2', `87.4', `93.2']]

\textbf{\textit{\textcolor{brown}{Question:}}} Please extract the named entity from the given table and output a JSON object that contains the following: \{`Task': [list of entities], `Dataset': [list of entities], `Model': [list of entities], `Method': [list of entities], `Metric': [list of entities], `Setting': [list of entities]\}. If no entities are presented in any categories keep it None.

\textbf{\textit{\textcolor{cyan}{Entities:}}} 
\{`Task': [], `Dataset': [`Dev', `Test', `SQuAD 1.1'], `Model': [`Human', `\#1 Ensemble-nlnet', `\#2 Ensemble - QANet', `BiDAF+ELMo (Single)', `R.M. Reader (Ensemble)', `bertbase', `bertlarge', `bertlarge', `bertlarge', `bertlarge', `BERT ensemble'], `Method': [], `Metric': [`EM', `F1', `EM', `F1'], `Setting': [`Top Leaderboard Systems (Dec 10th, 2018)', `Published', `Ours', `Single', `Single', `Ensemble', `Sgl.+TriviaQA', `Ens.+TriviaQA']\} \\\\\\

\textbf{\textit{\textcolor{purple}{Given type set:}}} [`Task', `Model', `Method, `Dataset', `Metric', `Setting']. 

\textbf{\textit{\textcolor{red}{Table:}}} \textcolor{red}{[T]} 

\textbf{\textit{\textcolor{brown}{Question:}}} Please extract the named entity from the given table and output a JSON object that contains the following: \{`Task': [list of entities], `Dataset': [list of entities], `Model': [list of entities], `Method': [list of entities], `Metric': [list of entities], `Setting': [list of entities]\}. If no entities are presented in any categories keep it None.

\textbf{\textit{\textcolor{cyan}{Entities:}}}  \\

\bottomrule
\caption{The prompt template of table NER leveraging 2-shot In-context Learning. \textcolor{red}{[T]}  denotes the table we want to extract.} 

\label{tab:tablener}
\end{tabular}
\end{table*}

\begin{table*}[!t]\footnotesize
\centering
\small
\label{table:table_RE_prompt}
\begin{tabular}{p{0.95\linewidth}}
\toprule
\textbf{\textit{Prompts of table RE (2-shot In-context Learning)}} \\ \midrule
\textbf{Considering relations between cells in tables: } \\
 \\ \\
\textbf{Here are two demonstrations:} \\\\\\

\textbf{\textit{\textcolor{red}{Table:}}} 
[[`System', `MNLI-(m/mm)', `QQP', `QNLI', `SST-2', `CoLA', `STS-B', `MRPC', `RTE', `Average'], 
[`', `392k', `363k', `108k', `67k', `8.5k', `5.7k', `3.5k', `2.5k', `-'], 
[`Pre-OpenAI SOTA', `80.6/80.1', `66.1', `82.3', `93.2', `35.0', `81.0', `86.0', `61.7', `74.0'], [`BiLSTM+ELMo+Attn', `76.4/76.1', `64.8', `79.8', `90.4', `36.0', `73.3', `84.9', `56.8', `71.0'], [`OpenAI GPT', `82.1/81.4', `70.3', `87.4', `91.3', `45.4', `80.0', `82.3', `56.0', `75.1'],
[`bertbase', `84.6/83.4', `71.2', `90.5', `93.5', `52.1', `85.8', `88.9', `66.4', `79.6'], 
[`bertlarge', `86.7/85.9', `72.1', `92.7', `94.9', `60.5', `86.5', `89.3', `70.1', `82.1']]

\textbf{\textit{\textcolor{brown}{Question:}}} Please extract all relations from the given table and output a JSON object that contains the following: \{[cell: cell]\}. If no relations are presented keep it None.

\textbf{\textit{\textcolor{cyan}{Relations:}}} 
\{[`Pre-OpenAI SOTA':`MNLI-(m/mm)', `Pre-OpenAI SOTA':`QQP',`Pre-OpenAI SOTA':`QNLI',`Pre-OpenAI SOTA':`SST-2', 
`Pre-OpenAI SOTA':`CoLA', `Pre-OpenAI SOTA':`STS-B', `Pre-OpenAI SOTA':`MRPC', `Pre-OpenAI SOTA':`RTE', `Pre-OpenAI SOTA':`Average',
`Pre-OpenAI SOTA':`80.6/80.1',`Pre-OpenAI SOTA':`66.1',`Pre-OpenAI SOTA':`82.3',`Pre-OpenAI SOTA':`93.2',`Pre-OpenAI SOTA': `35.0',`Pre-OpenAI SOTA':`81.0',`Pre-OpenAI SOTA':`86.0',`Pre-OpenAI SOTA':`61.7',  `Pre-OpenAI SOTA':`74.0']\}  \\\\\\

\textbf{\textit{\textcolor{red}{Table:}}} 
[[`System', `Dev', `Dev', `Test', `Test'], [`', `EM', `F1', `EM', `F1'], [`Top Leaderboard Systems (Dec 10th, 2018)', `Top Leaderboard Systems (Dec 10th, 2018)', `Top Leaderboard Systems (Dec 10th, 2018)', `Top Leaderboard Systems (Dec 10th, 2018)', `Top Leaderboard Systems (Dec 10th, 2018)'], [`Human', `-', `-', `82.3', `91.2'], [`\#1 Ensemble - nlnet', `-', `-', `86.0', `91.7'], [`\#2 Ensemble - QANet', `-', `-', `84.5', "90.5'], [`Published', `Published', `Published', `Published', `Published'], [`BiDAF+ELMo (Single)', `-', "85.6', `-', `85.8'], [`R.M. Reader (Ensemble)', `81.2', `87.9', `82.3', `88.5'], [`Ours', `Ours', `Ours', `Ours', `Ours'], [`bertbase(Single)', `80.8', `88.5', `-', `-'], [`bertlarge(Single)', `84.1', `90.9', `-', `-'], [`bertlarge(Ensemble)', `85.8', `91.8', `-', `-'], [`bertlarge(Sgl.+TriviaQA)', `84.2', `91.1', `85.1', `91.8'], [`bertlarge(Ens.+TriviaQA)', `86.2', `92.2', `87.4', `93.2']]

\textbf{\textit{\textcolor{brown}{Question:}}} Please extract all relations from the given table and output a JSON object that contains the following: \{[cell: cell]\}. If no relations are presented keep it None.

\textbf{\textit{\textcolor{cyan}{Relations:}}} 
\{[`\#1 Ensemble - nlnet':`Dev', `\#1 Ensemble - nlnet':`EM', `\#1 Ensemble - nlnet': `Top Leaderboard Systems (Dec 10th, 2018)', `\#1 Ensemble - nlnet': `86.0', `\#1 Ensemble - nlnet': `Test',`\#1 Ensemble - nlnet':`F1', `\#1 Ensemble - nlnet': `91.7' ]\} \\\\\\

\textbf{\textit{\textcolor{red}{Table:}}} \textcolor{red}{[T]} 

\textbf{\textit{\textcolor{brown}{Question:}}} Please extract all relations from the given table and n output a JSON object that contains the following: \{[cell: cell]\}. If no relations are presented keep it None.

\textbf{\textit{\textcolor{cyan}{Entities:}}}  \\

\bottomrule
\caption{The prompt template of table RE leveraging 2-shot In-context Learning. \textcolor{red}{[T]} denotes the table we want to extract. Since the relations are very dense in tables, we here list part of the relations.}

\label{tab:tablere}
\end{tabular}
\end{table*}   

\end{document}